\documentclass[runningheads]{llncs}
\usepackage[T1]{fontenc}
\usepackage{amsmath} 
\usepackage{amssymb}  
\usepackage{acronym}
\usepackage{xcolor}
\usepackage{textcomp}
\usepackage{xspace}
\usepackage{multirow}      
\usepackage{graphicx}
\usepackage{subcaption}
\usepackage{booktabs}     
\usepackage{hyperref}
\usepackage{capt-of}  
\usepackage{booktabs}
\usepackage{multirow}
\usepackage{comment}
\usepackage{booktabs}  
\usepackage{amssymb}   
\usepackage{float}

\usepackage{graphicx}
\begin{document}
\title{An LLM-Based Automatic Sportscast Solution for Robot Soccer Matches}
%
%
\author{
F. Petri \inst{1,2,\dag} \thanks{\dag The authors contributed equally}
\orcidID{0009-0008-6208-1498}
\and
M. Brienza \inst{1,\dag}  \orcidID{0009-0000-1549-9500}
\and \\
D. Nardi\inst{1}\orcidID{0000-0001-6606-200X}
\and
D. D. Bloisi \inst{3}\orcidID{0000-0003-0339-8651}
\and \\
A. Gangemi \inst{2,4}\orcidID{0000-0001-5568-2684}
\and
V. Suriani \inst{1} \orcidID{0000-0003-1199-8358}
}
\authorrunning{Petri, Brienza et al.}
%
\institute{
Dept. of Computer, Control, and Management Engineering\\ Sapienza University of Rome, Rome (Italy),
    \email{francesco.petri@uniroma1.it}
    \email{\{lastname\}@diag.uniroma1.it}
\and
Institute for Cognitive Sciences and Technologies (ISTC-CNR), National Research Council, Italy
\and
Dept. of International Humanities and Social Sciences, 
International University of Rome, Rome (Italy),
\email{domenico.bloisi@unint.eu}
\and
University of Bologna, Bologna (Italy)
}
\maketitle              
\begin{abstract}
RoboCup has always been a scenario to develop systems that solve real-world problems. Driven by the main goal of playing against the 2050 FIFA World Cup champions, the RoboCup Soccer leagues need to constantly measure how the research community is progressing. Computing visual statistics from match videos is a crucial way to track this evolution. To address this challenge, this paper introduces a fully autonomous, real-time sports commentator for RoboCup matches. By bridging the gap between raw kinematic tracking and natural language generation, our neuro-symbolic architecture extracts precise statistics from video streams and turns them into fluent, hallucination-free narration. The proposed system is capable of generating statistics and commentary both during live match streaming and in post-game analysis, easily adapting to the new dynamism of the league where different humanoid robots of different sizes share the field. Supplemental materials are available at \url{https://lab-rococo-sapienza.github.io/MARIO/}

\end{abstract}
\section{Introduction}

Since 1997, RoboCup's primary objective has been to promote scientific research and accelerate progress in artificial intelligence and robotics. 
These technologies can then be transferred to real-world problems ranging from household assistance to rescue operations and robotic soccer. Testing algorithms and new approaches in the dynamic environment of robotic soccer guarantees the accelerated development of real-time, embodied, and adaptive systems. However, an aspect that is often considered secondary with respect to technical and research challenges, but is becoming increasingly important, is the explainability and interpretability of the progress achieved by the community. In view of the 2050 objective of challenging the human FIFA World Cup champions, RoboCup Soccer needs mechanisms to observe, measure, and trace improvements over time. 
In this perspective, visual statistics and automatic commentary can play a central role: they transform raw match dynamics into understandable descriptions, allowing researchers to monitor the evolution of the league and enabling the general public to access meaningful explanations of the underlying scientific advances. 

\begin{figure}[t]
    \centering
    \includegraphics[width=0.8\linewidth]{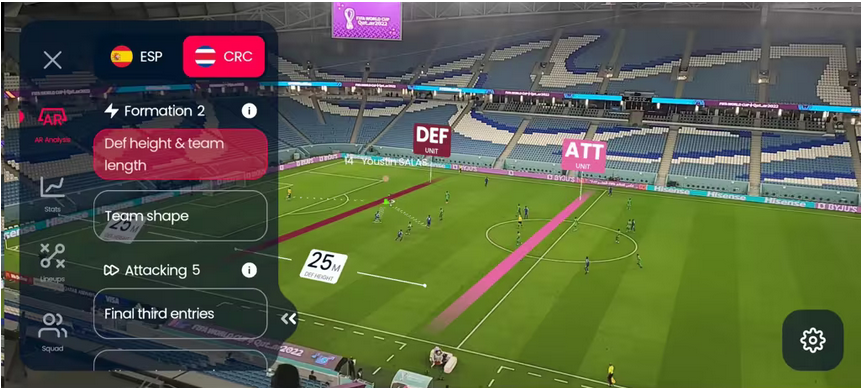}
    \caption{An example of the FIFA+ Stadium Experience used during the 2022 FIFA World Cup in Qatar. The application allows fans to overlay real-time statistics and heatmaps onto the pitch using augmented reality. Image from the official FIFA website \cite{fifa_stadium_exp_2022}.}
    \label{fig:fifa}
\end{figure}

A first attempt to solve this problem is represented by the Open Research Challenge introduced in the Standard Platform League (SPL) in 2022 \cite{RoboCupTechnicalCommittee2022}, which aimed to address the issue of obtaining statistics from matches similar to those in human soccer like, as happened, for example, during the 2022 World Cup (figure \ref{fig:fifa}). 
This challenge shifted the focus toward computer-vision-based systems capable of tracking robots and the ball directly from match videos. Considering also the contributions from the Small Size League (SSL), which already developed a system capable of generating detailed statistics from their accurate oracle-like sensor setup,\footnote{\url{https://github.com/RoboCup-SSL/ssl-match-stats}} we propose an analysis framework that can generate statistical data to track the future advancement of the nascent Humanoid Soccer League (HSL), which was first established this year and is certainly going to grow and evolve rapidly in the future.
Starting from our previous work, which successfully addressed the complex requirements of that challenge, we present an extension of the MARIO \cite{bloisi2022mario} (Modular and Extensible Architecture for Computing Visual Statistics) framework. While the original system established a robust foundation for visual perception and data extraction, combining MARIO's deterministic tracking with a symbolic abstraction of specific game events, this work enables LLMs to generate real-time, context-aware automated narration.
Automatic sportscasting is worth studying in robot soccer because it combines multiple RoboCup-relevant technical challenges \cite{hitoshi11999missionpaper} with the need to translate objective match data into interpretable, human-readable explanations.

While LLMs possess powerful capabilities for generating text following prompt instructions, a significant challenge arises when these models are tasked with processing raw numerical data \cite{fang2024large}. In our scenario, the visual system provides continuous streams of raw coordinates for the field, ball, and robots. Interpreting these data through prompting alone is unreliable: LLMs may hallucinate facts, misidentify the active team, or fail to capture spatial relationships. Although Vision Language Models (VLMs) and video foundation models can describe actions directly from images \cite{brienza2024llcoach}, they are not suitable for our setting. They do not yet meet the real-time requirements of live matches and lack persistent tracking mechanisms for producing consistent match summaries beyond frame-by-frame interpretation.

Our work extends the MARIO architecture with a neuro-symbolic approach that bridges the gap between noisy, low-level tracking signals and natural language generation. The system builds on a computer-vision tracking pipeline where neural networks localize robots and the ball; homography then maps image coordinates to field coordinates, yielding a plan-view trace of the match consistent with official field dimensions. In response to the soccer league’s evolution, we updated our perception module from a specific SPL setting to a more general soccer tracker capable of handling different field sizes and detecting various types of humanoid robots. Based on these data, we add a symbolic layer that employs rules, aggregations, and state detection to identify discrete sporting concepts, such as types of kicks, active teams, and the current game phase. Then, an LLM transforms these deterministic outputs into prototypal commentary which, by being grounded in explicit symbolic events, significantly reduces the risk of hallucinations.
To summarize, the contribution of this work is threefold:
\begin{enumerate}
    \item A symbolic event extraction module for RoboCup soccer videos, where game events are inferred from tracked dynamics and rule-based constraints.

    \item A sportscast policy mechanism for natural-language commentary generation, including priority scheduling, temporal gating, and state-aware controls to reduce hallucinations, and ensure coherence with RoboCup HSL game rules.
    
    \item A semi-automatic labeling tool for domain adaptation of lightweight CNNs for robot/player and team recognition with limited human supervision.
    
\end{enumerate}

The remainder of the paper is organized as follows. Section \ref{sec:rel_work} reviews related work, Section \ref{sec:methodology} presents the proposed approach, Section \ref{sec:results} reports the experimental results, and Section \ref{sec:conclusion} concludes the paper.

\section{Related work}
\label{sec:rel_work}

Sport analytics for soccer and other sports have been gaining increasing attention from the research community in recent years, whether as a context to perform classic tasks such as action recognition or decision making, or for practical domain-specific purposes such as producing or improving an automatic referee system \cite{DEOLIVEIRA2023100232} \cite{pu2024}. Morra et al. \cite{MORRA2020100612} design an event recognition framework for soccer, but it is limited to a simulated environment. Majeed et al. \cite{majeed2025real} propose a visual analysis system based on graph neural networks that works in real time, but being based on human soccer, they have access to a wealth of labeled data that is not available in the RoboCup context.

While most works focus on human sports, some analytics work has been done specifically in the RoboCup environment \cite{abreu2012} \cite{fukushima2020}. However, they rarely focus on humanoid robots, and never on visual analysis, instead obtaining their data from simulation logs or precise sensors mounted on the field. We employ our system on games with humanoid robots, as observed by a camera placed at an arbitrary position on the side of the field outside of our control.
Automatically producing sports commentary is a challenging problem in current literature due to needing to join multimodal processing, real-time constraints, and last but not least, a wish to entertain the audience \cite{Zheng2025FromMP}. LLMs have been found to be effective for the task, provided that they are supported by prompt engineering and game-specific data, as exemplified by Sun et al. in the case of basketball \cite{sun2025}.
Inspired by their work, we design a similar set of neurosymbolic events to ground the generated commentary, but applied to the specific rules and actions of soccer and RoboCup in particular rather than basketball. To the best of our knowledge, no such event-driven, real-time system exists for soccer, in particular in the RoboCup context. 

We also note that while multiple labeled datasets for human soccer and other sports exist \cite{Cioppa_2022} \cite{rao-etal-2024-matchtime} \cite{zhang-eickhoff-2021-soccer}, data featuring robots is significantly more limited. Action recognition models, such as those recently developed for basketball \cite{sun2025}, can significantly enhance the capabilities of visual tracking systems. Specifically, incorporating pose detection allows the system to extract a deeper semantic understanding of the robot's physical states. This introduces the ability to verify complex contextual conditions during the game, such as detecting when a robot has fallen \cite{zampino2022fall}.
\begin{figure}[t]
    \centering
    \includegraphics[width=\linewidth]{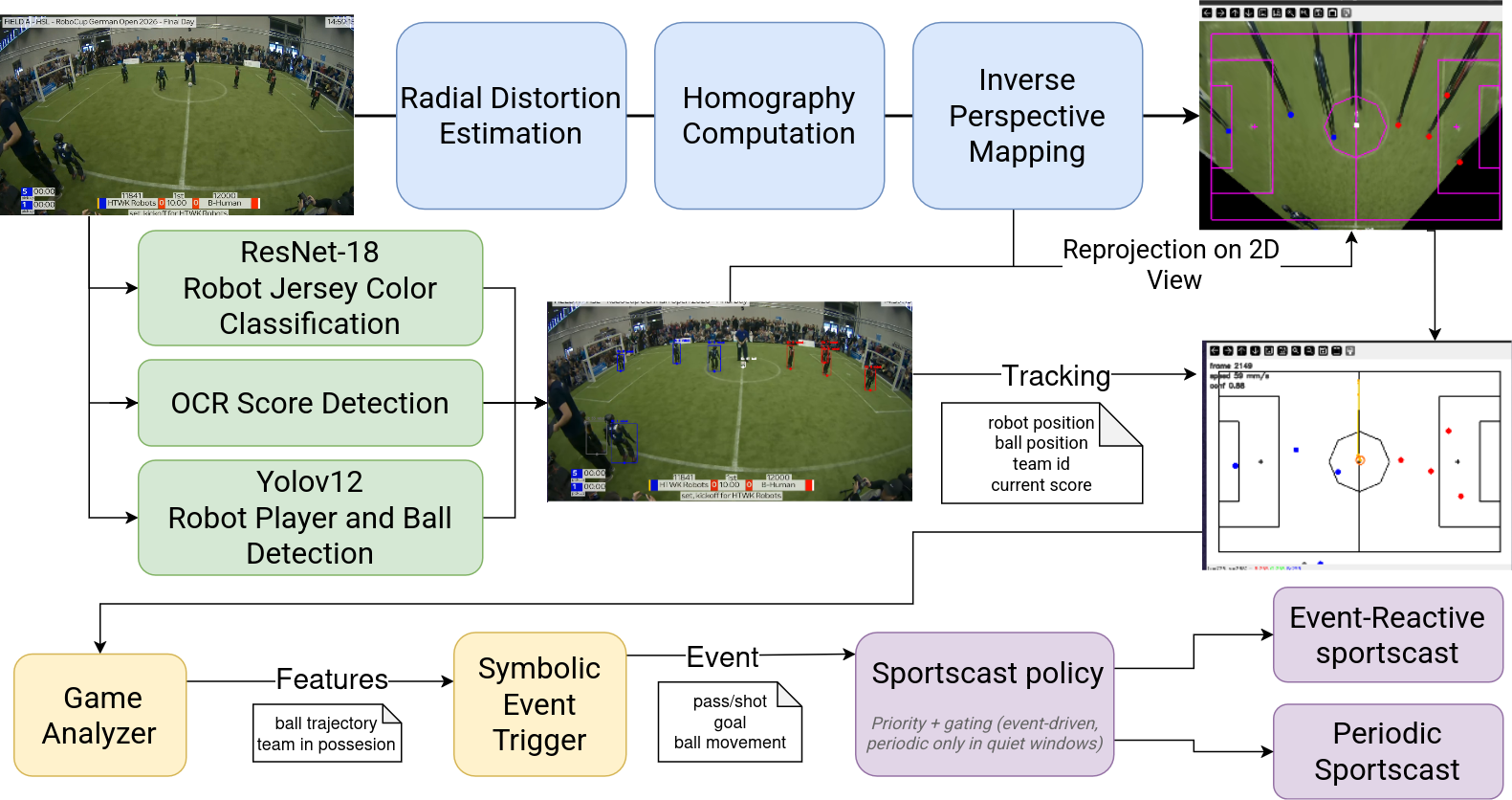}
    \caption{System architecture detailing the flow from raw visual perception and 2D field reprojection to symbolic event extraction and dual-role LLM-driven sportscasting.}
    \label{fig:architecture}
\end{figure}
\section{Methodology}
\label{sec:methodology}

This section details the architecture shown in Figure \ref{fig:architecture}. The methodology is organized around three core components: (i) a calibration module that estimates distortion and homography to map detected robots and the ball onto an undistorted 2D field representation, (ii) a symbolic event extractor that converts kinematic and scoreboard evidence into discrete soccer events, and (iii) a rule-aware sportscast policy with two complementary autonomy levels (event-reactive and periodic) for robust, coherent natural-language commentary.

\subsection{Calibrator Module}
\label{subsec:calibrator}

The calibrator maps pixels in the broadcast view to points on the RoboCup pitch (planar coordinates in millimetres). An interactive graphical interface allows the user to align landmarks in a source frame with a synthetic top-down template derived from official field markings (touchlines, halfway line, centre circle, goal and penalty areas, penalty marks) for the purpose of defining an homography, and to estimate the radial distortion model of the camera.

The radial distortion is defined by the following formula, calculated up to the third order \cite{Brown1971CloseRangeCC}:

$$
\begin{aligned}
x' &= x + \bar{x}(k_1 r^2 + k_2 r^4 + k_3 r^6) \\
y' &= y + \bar{y}(k_1 r^2 + k_2 r^4 + k_3 r^6)
\end{aligned}
$$
where $(\bar{x}, \bar{y})$ is the center of the transformation, and $r^2 = (x - \bar{x})^2 +  (y - \bar{y})^2$.

We make use of the exact inversion formulae presented in \cite{drap2016invrad} to enable two-way conversion between image space and field space.

For estimating the homography from the given points, we use a DLT algorithm with RANSAC as implemented by the opencv library\footnote{\url{https://docs.opencv.org/4.x/d9/dab/tutorial_homography.html\#tutorial_homography_Demo1}}.

\begin{figure}
    \centering
    \includegraphics[width=\linewidth]{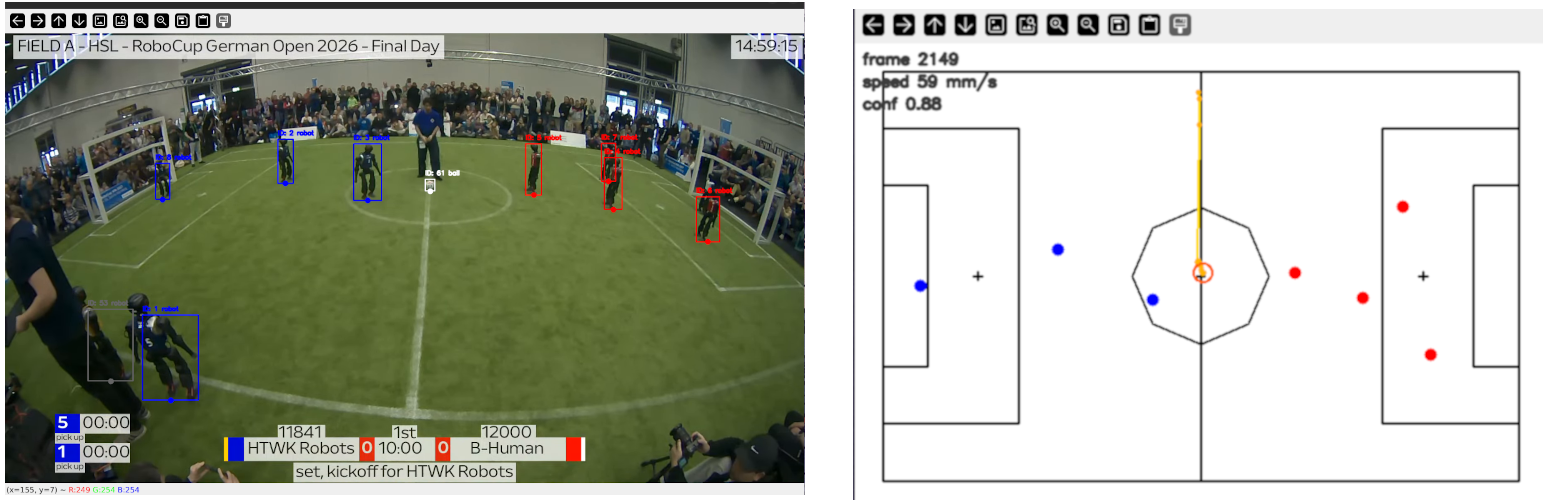}
    \caption{2D Reprojection of Visual Detections}
    \label{fig:mario}
\end{figure}

\subsection{Visual Perception Module}
\label{subsec:visual-perception}

This module converts each raw frame into a structured match representation: robot detections with persistent identities and team labels, ball position in field coordinates, and OCR scoreboard values. All outputs share the same metric field frame as the calibrator, so later modules operate on a unified geometric reference.

Online inference combines a fast detector (YOLOv12 \cite{tian2025yolov12}) for players and a lightweight jersey-color classifier (ResNet-18 \cite{he2016deep}) for team attribution. In parallel, an OCR branch reads left/right score digits and provides explicit game-state supervision for goal confirmation.
The choice is based on the assumption of only having access to public streaming, though GameController data can be used as an alternative source if it becomes available.
An example of the visual perception module in action is depicted in figure \ref{fig:mario}, showing the bounding boxes for robot and ball detections onto the video as well as a view of the corresponding reprojection in field space. The latter also shows some basic stats, such as the ball velocity.

\subsubsection{Robot Player Detection}

To preserve real-time throughput while adapting to the evolving morphology of Humanoid Soccer League platforms, we use a semi-automatic data adaptation workflow. Heavy foundation models are used offline for annotation bootstrapping, while compact models are deployed online for low-latency inference. Candidate bounding boxes for humanoid robots and the ball are generated from match frames via SAM 3 \cite{carion2025sam}. Robot crops are then extracted and queried with a Vision-Language Model (VLM) to label jersey colors. An optional human-in-the-loop step refines boxes and color labels before training. The resulting dataset is used to train YOLOv12 for detection and tracking support, and ResNet-18 for jersey color recognition.

\subsection{Sportscast Policy}
At any frame $t$, the module takes the tracked robots $X_t = \{(id_i, team_i, x_i, y_i)\}$, 
the ball position $b_t = (x^b_t, y^b_t)$, and the OCR score $Q_t = (q^L_t, q^R_t)$. To 
evaluate actions, we compute the effective ball displacement $d_{\text{eff}}$
relative to its last known position and velocity $v_{\text{eff}}$ as 
$d_{\text{eff}} / \Delta t$, where $\Delta t$ is the elapsed time since that last known 
position. If $\Delta t$ exceeds a predefined threshold, these quantities are considered 
unreliable and only the ball's current position is used instead.
To reduce tracking noise, these quantities are calculated as the maximum between a short-window estimate and a trajectory trace. Game hints are computed based on proximity and direction:
\begin{itemize}
    \item \textbf{Possession Hint:} We identify the closest robot $i^\star_t = \arg\min_i \lVert p_i - b_t\rVert_2$. If the distance $r_t = \lVert p_{i^\star_t} - b_t\rVert_2$ falls below a defined threshold $r_{\text{poss}}$, the ball is considered possessed by $team_{i^\star_t}$.
    \item \textbf{Directional Intent:} We compute the normalized ball trajectory vector $\hat{u}_t$. By calculating the dot products (directional cosines) against the vector pointing towards a teammate ($\hat{u}_{\text{team}}$) and the vector pointing towards the opponent's goal ($\hat{u}_{\text{goal}}$), we obtain $c_{\text{team}}$ and $c_{\text{goal}}$. These values indicate whether the ball is moving intentionally toward a friendly player or a target goal.
\end{itemize}

\subsubsection{Event Triggering Logic}
The hints are passed through logical functions to detect discrete events. An action is first validated by checking if the ball has sufficient kinematic energy ($v_{\text{eff}} \ge v_{\min}$, $d_{\text{eff}} \ge d_{\min}$), filtering out instances where the ball is manually repositioned by a referee or movements too small to be considered significant events.
Valid actions are then classified using predefined directional and distance thresholds. An action is categorized as a pass if the ball's trajectory aligns closely enough with the position of a teammate. Conversely, an action is classified as a shot if it meets three conditions: it does not strongly align with a teammate (ruling out a pass), its trajectory is directed accurately toward the opponent's goal, and the ball travels a minimum required distance forward during the play (to avoid rebounds).
Additionally, definitive goal events are strictly triggered by a positive difference in the OCR-read scoreboard state ($q^L_t - q^L_{t-1}$ or $q^R_t - q^R_{t-1}$). The resulting output is a structured event tuple containing the action class, the acting team, and the associated confidence.
Translating raw events directly into speech often results in repetitive, unnatural, or hallucinated commentary. To bridge the gap between robotic perception and human-like broadcasting, the \emph{Sportscast Policy} module routes the validated events to a Large Language Model (LLM) using a strict priority and gating system, operating on two complementary autonomy levels: Event-Reactive and Periodic.

\subsubsection{Event-Reactive Commentary and Gating}
Event-reactive commentary is activated when a validated, high-priority event is emitted by the symbolic layer. To ensure that urgent occurrences preempt lower-priority queued outputs, emission follows a strict priority hierarchy, ranking goals above shots or passes, which in turn outrank other minor events and periodic generation. To avoid repetitive or unstable narration, the policy applies several temporal guardrails before executing any call to the Large Language Model.
These guardrails, implemented as explicit, algorithmic rules in the source code, suppress duplicate events, enforce a cooldown between closely timed events of similar priority by keeping only the earlier one, suppress events with a low-priority, and discard stale data, such as outdated ball positions or scores, once it exceeds a freshness threshold, until new detections become available.
If an event successfully passes all these gates, the system constructs a conditioned prompt using machine-validated fields, such as the event class, acting team, and current score. The generation contract strictly restricts the model to producing one short, factual sentence without introducing unsupported entities or actions. Consequently, the core event semantics remain strictly determined by the symbolic logic, while the language model is solely utilized for natural linguistic realization.

\subsubsection{Periodic Commentary for Engagement}
Conversely, periodic commentary is enabled exclusively during quiet windows, which occur when no recent high-priority events are active and no guardrail would prevent the activation of a new event.
Unlike the event-reactive approach, this branch is conditioned on aggregated scene context rather than a single discrete trigger. The contextual inputs include the current score state, short-term possession trends, the spatial zone of the ball, recently extracted symbolic events, and a short memory buffer of spoken history to create context to the sportscast having a flow during the comments.
The periodic prompt instructs the model to generate a single, concise scene-level sentence aimed at maintaining narrative continuity and audience engagement.
This generation is subject to the same strict grounding constraints as the reactive module, explicitly preventing the system from hallucinating unverified game phases such as the kick off where the LLM have robots standing still without the ball and could still infer phases of indecision in the game.

\section{Experimental Results}
\label{sec:results}

Experimental results are validated both quantitatively and qualitatively.
Quantitative validation is performed by testing our system on three video clips that were streamed on YouTube on March 14th, 2026, during the RoboCup German Open 2026 testing the system in the new league with new robot and larger field. We use the clips as they are and process them causally, i.e. without looking at future frames, to simulate a real-time stream.
Qualitatively, we manually assess whether the generated sportscast comments are coherent with the observed game actions; occasional false positives are still observed in rare ball-tracking edge cases where false positive detection of the ball or movement from human referee of the ball happens, but this condition is reduced by trigger conditions and safety gates tuning parameters. 
We evaluate the accuracy of robot detection and tracking by calculating the Euclidean distance in field space between the positions of the robots detected in the video and the positions those same robots reported to the GameController (GC), i.e. the referee's central computer, during the game. GC data does not necessarily represent ground truth, since the robot's estimate of its own position is ultimately dependent on the quality of its localization. For this reason, we carried out a qualitative evaluation of the GC logs by superimposing the reported positions onto the video clips, and found that at least one team's localization is sufficiently accurate that it can be considered close enough to ground. Conversely, teams that failed to report their position to the GC at all were excluded from this evaluation. Robots send reports to the GC 1 or 2 times per second, i.e. once per 15-30 video frames. This makes the GC data sparser than the video, but we smooth this out by averaging the error over multiple frames, as we explain in the following.

As the evaluation metric, we calculate the frame-by-frame Root Mean Squared Error (RMSE) as follows: on each frame $f$, we consider the set $V_f$ of all robots that have been detected in the video as belonging to a team with good localization, i.e. wearing the corresponding jersey color, and the set $L_f$ of all positions from the GC logs of robots of the same team that were playing (i.e. not penalized) at that time. Then, for each element of the larger set, we compute the Euclidean distance to the closest element of the smaller set, and take the average. Distinguishing the larger and the smaller set is important because it naturally penalizes detecting more or less robots than what were logged to have been actually playing: in both cases, more distances are calculated and more error is accumulated.
\begin{equation}
    RMSE_f = 
    \begin{cases} 
    \dfrac{1}{|V_f|} \sum_{p \in V_f} \min_{q \in L_f} \|p - q\|_2 & \text{if } |V_f| \geq |L_f| \\[1.2em]
    \dfrac{1}{|L_f|} \sum_{p \in L_f} \min_{q \in V_f} \|p - q\|_2 & \text{if } |V_f| < |L_f|
    \end{cases}
\end{equation}
\begin{table}[t]
    \centering
    \setlength{\tabcolsep}{6pt}    
    \begin{tabular}{lccc}
        \toprule
        \textbf{Clip name} & \textbf{Mean} & \textbf{Standard deviation} & \textbf{(25\%, 50\%, 75\%) quartiles} \\
        \midrule
        Middle 1st & 0.637 & 0.551 & (0.207, 0.446, 0.848) \\
        Middle 2nd & 0.657 & 0.377 & (0.394, 0.513, 0.812) \\
        Large 1st  & 0.930 & 0.795 & (0.339, 0.610, 1.247) \\
        \bottomrule
    \end{tabular}
    \caption{Aggregated RMSE for our video clips, with entries measured in meters.}
    \label{tab:results-aggr}
\end{table}
\begin{figure}[t]
    \centering

    \begin{subfigure}[b]{0.49\textwidth}
        \centering
        \includegraphics[width=\textwidth, trim={0.4cm 0cm 1.6cm 0.8cm}, clip]{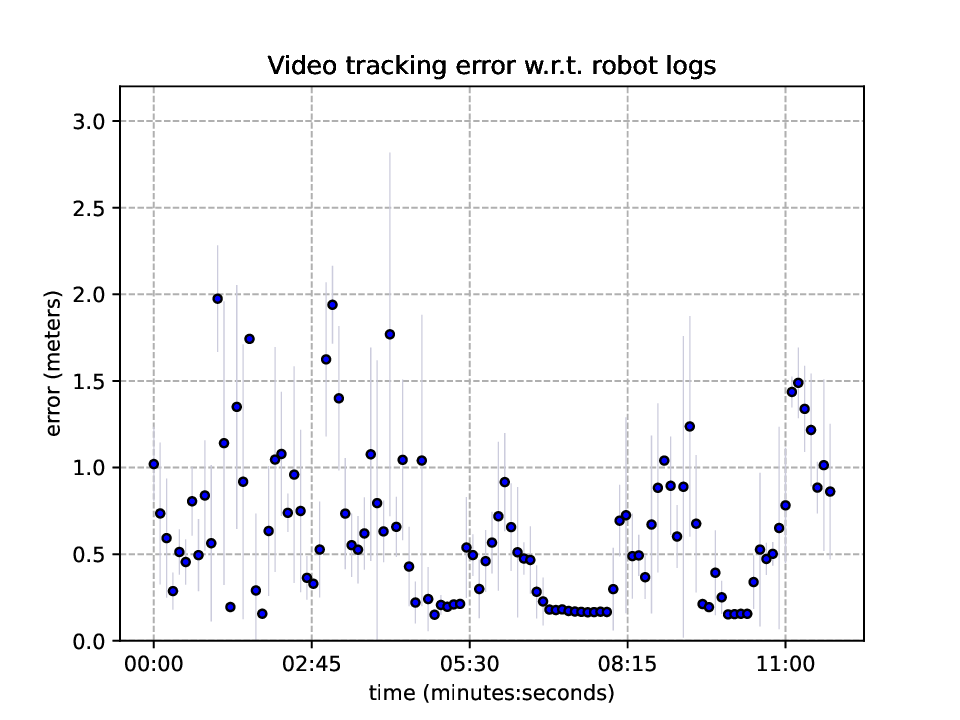}
        \caption{``Middle 1st'' clip.}
        \label{fig:error-m1}
    \end{subfigure}
    \hfill
    \begin{subfigure}[b]{0.49\textwidth}
        \centering
        \includegraphics[width=\textwidth, trim={0.4cm 0cm 1.6cm 0.8cm}, clip]{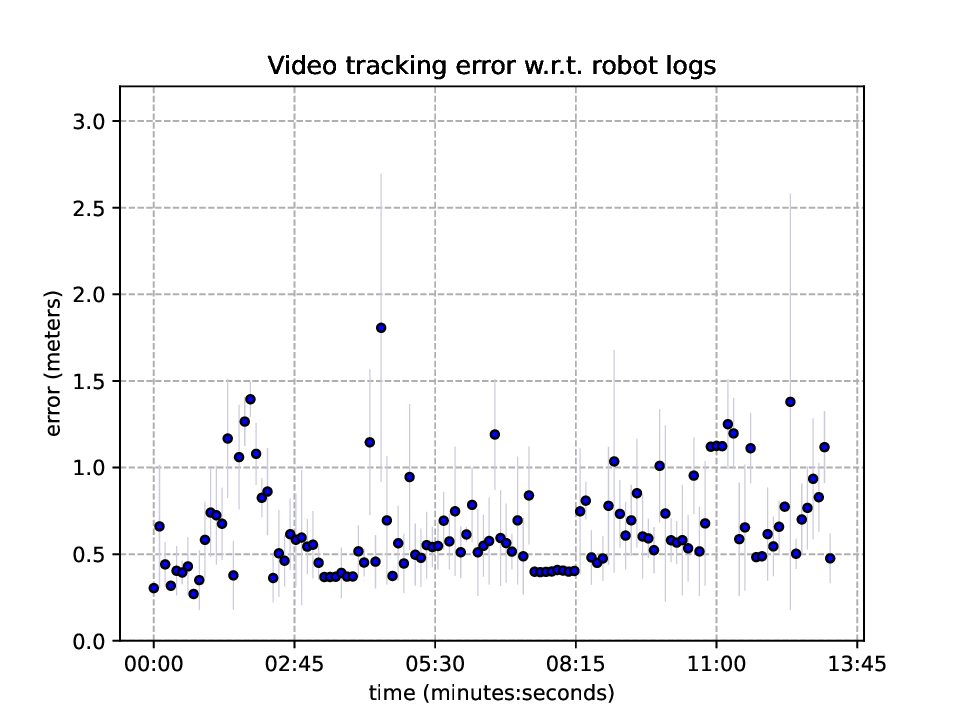}
        \caption{``Middle 2nd'' clip.}
        \label{fig:error-m2}
    \end{subfigure}

    \vspace{0.5em}

    \begin{subfigure}[b]{\textwidth}
        \centering
        \includegraphics[width=0.49\textwidth, trim={0.4cm 0cm 1.6cm 0.8cm}, clip]{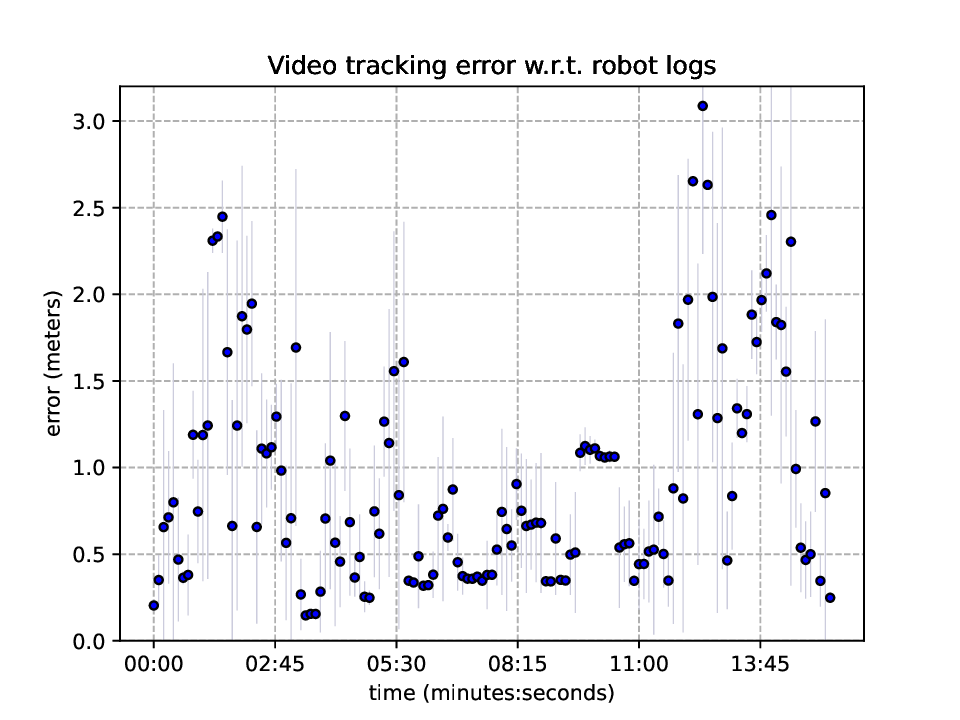}
        \caption{``Large 1st'' clip.}
        \label{fig:error-l1}
    \end{subfigure}

    \caption{RMSE over time for all video clips. Each point represents the average of 200 frames. Some error bars have been cut off to preserve the presentation of the data points, but no bar reaches higher than 4.4 meters.}
    \label{fig:combined-results}
\end{figure}
We calculate the error throughout the whole length of the clips. Aggregated results are presented in table \ref{tab:results-aggr}. We can observe that the tracking error is below $1.3\ m$ at least $75\%$ of the time. Considering that all clips were recorded in a $14 \times 9\ m$ field, and that generating sports commentary is based on describing a dynamic situation in terms of rough attack and defense rather than precisely localizing each player, we consider this sufficiently accurate for the purposes of automatic sportscast generation.

We also report the results in figure \ref{fig:combined-results}. To avoid visual cluttering of the data points, we subdivide the video in contiguous, non-overlapping 200-frame windows and represent the mean within each window with one data point. Error bars for the standard deviation have been added to represent the variance within each window. As a positive side effect, the error fluctuation that would result from comparing time-dense video detections with time-sparse logs is smoothed out.
We note that the error appears to have wide fluctuations, as expected of the high dynamicity of the environment and the lack of perfect ground truth data, but still, the spikes in error represent infrequent outliers, as evidenced by the quartile analysis above (table \ref{tab:results-aggr}). We also see spans of time where the error stabilizes: these correspond to moments where the robots stop playing, for example before kickoff or during a referee decision.
\section{Conclusion and Future Work}
\label{sec:conclusion}

This work introduced a novel, fully automated sportscasting framework for the RoboCup Soccer League. By combining a robust visual perception and tracking pipeline with a rule-aware symbolic event trigger and a dual-role LLM policy, we demonstrated the feasibility of generating coherent, real-time commentary directly from raw video feeds, also during live streaming enabling a new frontier of people engament in RoboCup.

A key advantage of this methodology is its structural adaptability. Because the system is designed as a dynamic, data-driven pipeline, it is not strictly bound to a single robot morphology and can be readily scaled to serve all RoboCup leagues.
Additionally, the integration of Large Language Models naturally unlocks real-time multilingual capabilities, allowing the commentary to be dynamically translated and streamed directly to global audiences via platforms like YouTube, as a future goal of the league.
\begin{credits}
\subsubsection{\ackname} This work has been funded by the Italian National PhD Program on Artificial Intelligence run by Sapienza University of Rome in collaboration with the Italian National Council for Research, and by the Italian PNRR MUR project PE0000013-FAIR. Michele Brienza is funded by the European Union - Next Generation EU, Mission I.4.1 Borse PNRR Pubblica Amministrazione (Missione 4) Component 1 CUP B53C23003540006.
\subsubsection{\discintname}
The authors have no competing interests to declare that are
relevant to the content of this article.
\end{credits}
%
%
%
%
\bibliographystyle{splncs04}
\bibliography{biblio}
\end{document}